\documentclass{article}

\usepackage{arxiv}

\usepackage[utf8]{inputenc} 
\usepackage[T1]{fontenc}    
\usepackage{hyperref}       
\usepackage{url}            
\usepackage{booktabs}       
\usepackage{amsfonts}       
\usepackage{nicefrac}       
\usepackage{microtype}      
\usepackage{lipsum}		
\usepackage{graphicx}
\usepackage{natbib}
\usepackage{doi}
\usepackage{amssymb}
\usepackage{amsmath}
\usepackage{bbm}
\usepackage{multirow}
\usepackage[english]{babel}
\usepackage{graphicx}
\usepackage{tabularx}
\usepackage{placeins}

\title{Integrating Transformations in Probabilistic Circuits}

\author{Tom Schierenbeck \thanks{Correspondence to Tom Schierenbeck, \texttt{tom\textunderscore sch@uni-bremen.de}.} \\
	Institute for Artificial Intelligence\\
   University of Bremen\\
Am Fallturm 1, 28359 Bremen \\
\texttt{tom\textunderscore sch@uni-bremen.de} \\
\And
    	Vladimir Vutov \\
    Institute for Statistics\\
    University of Bremen\\
    Bibliothekstraße 1, 28359 Bremen \\
    \texttt{vkvutov@uni-bremen.de} \\
    \And
    	Thorsten Dickhaus \\
    Institute for Statistics\\
    University of Bremen\\
    Bibliothekstraße 1, 28359 Bremen \\
     \texttt{dickhaus@uni-bremen.de} \\
    \And
    Michael Beetz \\
    Institute for Artificial Intelligence\\
    University of Bremen\\
    Am Fallturm 1, 28359 Bremen \\
   \texttt{beetz@cs.uni-bremen.de}
}

\renewcommand{\headeright}{}
\renewcommand{\undertitle}{}
\renewcommand{\shorttitle}{Schierenbeck et al.: Integrating Transformations in Probabilistic Circuits}

\hypersetup{
pdftitle={Integrating Transformations in Probabilistic Circuits},
pdfsubject={q-bio.NC, q-bio.QM},
pdfauthor={Schierenbeck et al.},
}

\begin{document}
\maketitle

\begin{abstract}
This study addresses the predictive limitation of probabilistic circuits and introduces transformations as a remedy to overcome it. We demonstrate this limitation in robotic scenarios. 
We motivate that independent component analysis is a sound tool to preserve the independence properties of probabilistic circuits. Our approach is an extension of joint probability trees, which are model-free deterministic circuits. By doing so, it is demonstrated that the proposed approach is able to achieve higher likelihoods while using fewer parameters compared to the joint probability trees on seven benchmark data sets as well as on real robot data. Furthermore, we discuss how to integrate transformations into tree-based learning routines. Finally, we argue that exact inference with transformed quantile parameterized distributions is not tractable. However, our approach allows for efficient sampling and approximate inference.
\end{abstract}


\section{Introduction}
\label{sec:intro}

Joint probability distributions play an important role in uncertainty and artificial intelligence (AI) by allowing us to model complex relationships between multiple variables. These distributions provide an efficient representation of probabilistic knowledge and can lead to improved predictive accuracy compared to modeling variables independently. With joint probability distributions, it is possible to handle missing data and make predictions even when some information is not available. Additionally, joint probability distributions are scalable and can be used to handle large amounts of data, making them well-suited for big data applications. Using joint probability distributions in terms of AI allows us to more accurately represent and analyze complex data, making it a valuable tool in many AI-related tasks.

Especially in robotics, the use of joint probability distributions is important due to the complex and dynamic nature of the domain. In robotics, it is necessary to model and reason about relationships between multiple variables, such as the relationship between an object's and a robot's position or between an object's appearance and its identity. Joint probability distributions provide an efficient and flexible way to represent such relationships, allowing robots to make informed decisions based on uncertain and changing information. However, the current state-of-the-art probabilistic modeling struggles to model simple relations common in robotics. 

Imagine a robot gets an order to grasp different objects. Due to the physical restrictions of the robot, it always has to go to the upper right position relative to the object it wants to grab. Sampling from an experiment designed this way results in four-dimensional observations where the robot's and objects' positions in two dimensions are recorded.
These two positions are strongly dependent, yet this dependence can be expressed easily by linear relationships such as $X_{Robot} - X_{Object} \sim \mathcal{U}(0,1)$ and $Y_{Robot} - Y_{Object} \sim \mathcal{U}(0,1)$. If probabilistic circuits (PCs) are fitted on the raw data, there are numerous leaves describing this behavior for every possible point. However, this behavior becomes imprecise to some degree since the leaf distributions in PCs assume independence. 

\begin{figure}
    \centering
    \includegraphics[width=\linewidth]{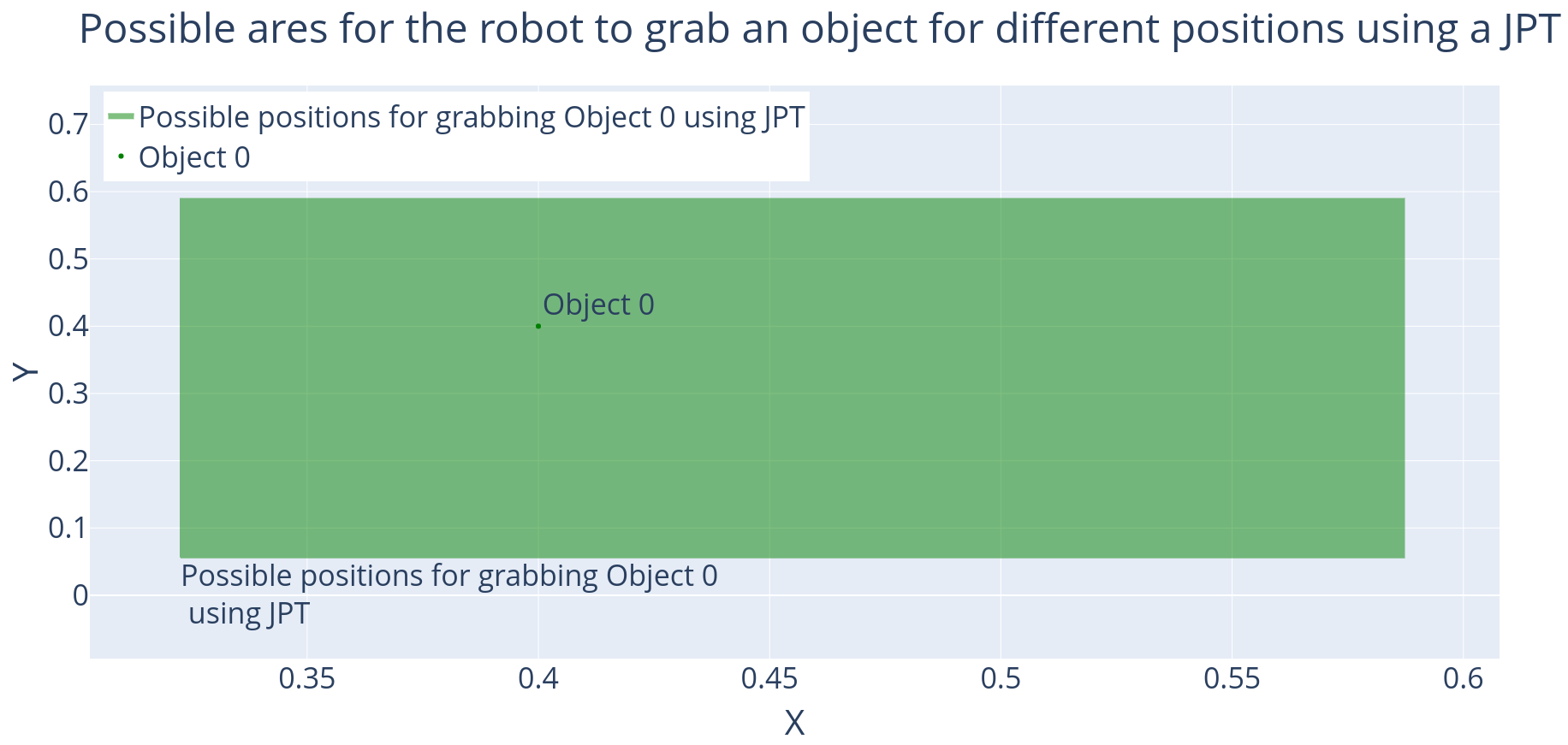}
    \caption{Possible area for the robot to grab an object modeled by a JPT with 104 parameters. The point represents an object. The green area represents all most likely positions predicted by the model.}
    \label{fig:jpt_predictions_grabbing}
\end{figure}

Figure \ref{fig:jpt_predictions_grabbing} depicts the predictions on where to stand to grab an object modeled by joint probability trees (JPTs) (see, \cite{nyga2022jpt}). As plotted, the possible area to stand in is not limited to the upper right corner of the object.
Hence, the representation is neither correct nor compact. Therefore, the expressiveness of JPTs and PCs, in general, needs to be improved such that direct influences can be modeled with an adequate amount of parameters. Furthermore, the dependencies of random variables are only rarely known in advance. Thus, the model should be able to extract dependencies from data without prior knowledge.

\subsection*{Limits of Probabilistic Circuits}
In recent years, PCs have contributed extensively to understanding and systematizing the landscape of joint probability distributions. Specifically,  \cite{choi2020probabilistic} showed that every tractable model is either a PC or a determinant point process. Furthermore, they introduce the shallow representation of a PC. The shallow circuit has the form
\begin{align}
\label{eq:prob_sum}
P(x_1, \dots ,x_m) = \sum_{l=1}^L w_l P_l(x_1, \dots ,x_m),\\
P_l(x_1, \dots ,x_m) = \prod_{j=1}^{m} P_{lj}(x_j) .\label{eq:fully_factorized}
\end{align}
Namely, it is a weighted sum of fully factorized distributions, where $w_l$ describes a weight for a leaf distribution $P_l$ and $P_{lj}(x_j)$ describes the distribution of variable $x_j$ in leaf $l$.  Equation (\ref{eq:fully_factorized}) is the definition of a fully factorized distribution.

In the context of PC that is tractable for most probable explanations (MPE), \cite{choi2020probabilistic} proved that this sum also has to be deterministic. Therefore, the PC can be considered a partitioning of the problem's domain, as \cite{nyga2022jpt} demonstrated.
Throughout this work, we use the terms "deterministic" and "determinism" in the sense of Definition 30 in \cite{choi2020probabilistic}.

Since any PC can be unrolled into its shallow representation, a limit for its predictive power is introduced. In a fully factorized distribution, variables are completely independent of each other; hence no predictive power is achieved. Therefore, the joint distribution is given by a product of the independent probability density functions (PDFs). If queries that predict something other than probability or likelihood alter the evidence slightly, the prediction is not (desirably) changed if the partition of the shallow representation is not changed.
Those queries are, for example, MPEs or queries that calculate the moment of a distribution.
Prediction in our context means to infer from a set of observed variables the evidence pertaining to other, unobserved variables.

Real-world applications barely involve analyses of (stochastically) independent variables, and such an assumption is too restrictive and hard to be validated (see, e.g., \cite{stange2016multiplicity}). On the contrary, nowadays, most studies examine data sets encompassing (many) dependent variables.


Multivariate statistical analysis (e.g., see in \cite{johnson2002applied}) offers a plausible way to decompose the original feature space into a (lower-dimensional) subspace with resulting independent vectors. In other words, we aim at yielding another basis that is a (linear) combination of the original (data) basis and that re-expresses the raw data.

In this work, we propose an extension of the JPT approach. Specifically, we do so under the scope of independent component analysis and by adopting the JPT methodology. We call our method \underline{i}ndependent \underline{c}omponent analysis - joint probability \underline{trees} ("IC-Trees").

The remainder of the paper is structured as follows. In Section \ref{sec:ica}, we briefly introduce the independent component analysis and its assumptions. In the following Section \ref{sec:ictrees}, we discuss the incorporation of the ICA method in the context of the JPTs framework. Section \ref{sec:experiments} is dedicated to the quantitative performance of our methodology based on several real data sets. Section \ref{sec:related_work} discusses related works, and we conclude with a discussion and outlook in Section \ref{sec:conclusion}.

\section{Independent Component Analysis}
\label{sec:ica}

Overcoming the limits of PCs cannot be done by adding more parameters since they only increase the resolution of the partitions. Instead of adding more parameters, we transform features into more meaningful variables. Nevertheless, these new variables should integrate well with the structure of probabilistic circuits. 

In the context of statistical data analysis, a classical method to carry out feature extraction and data reduction is constituted by principal component analysis (PCA). The latter approach is widely used in many data applications and provides a viable way to represent the original features by a smaller set of variables. PCA incorporates in its representation solely second-order moments of data distribution and not considering higher-order moments (statistics). In contrast, independent component analysis (ICA) is a generalization of PCA, focusing on higher-order moments and maximizing the independence of the transformed variables, which integrates well with the product nodes of PCs. Moreover, PCA can be employed as a pre-processing step in some ICA algorithms (for more details, \cite{bugli2007comparison}).

To formalize the ICA model, we assume a data set that is stored as a $n \times m$ matrix $X = (x_{ij})_{\substack{1 \leq i \leq n \\ 1 \leq j \leq m}}$, where $m$ denotes the number of features or covariates, and $n$ is the number of observational units. Further, in this study and without loss of generality, we assume there exist $m$ independent "latent variables" $s_{1}, \ldots, s_m$, such that $x_{j} = a_{1j}s_{1} + a_{2j}s_{2} + \cdots + a_{mj}s_{m}$, for all $j \in \{ 1, \ldots, m \}$, where $x_j$ is the random variable corresponding to the $j$-th column of $X$. Each such $x_j$ is thus modeled by a linear mixture of latent (i.e., unobserved) independent random variables. 

In matrix form, the ICA model is given by the equation (cf.\cite{hyvarinen2000independent, bugli2007comparison})
\begin{equation}
    \label{ICA_model_equation}
  x = As,
\end{equation}
where $s = \left(s_{1}, \ldots, s_{m} \right)^{T}$ is an $ m \times 1$ column vector gathering the $m$ random variables which describe the "source signals", and $x = \left(x_{1}, \ldots, x_{m} \right)^{T}$ represents the $m \times 1$ column vector of the $m$ observable signals. The "mixing matrix" $A$ (a square $m \times m$ matrix) consists of the (linear) mixture coefficients.

Following estimating the matrix $A$, one can calculate its inverse, denoted by $W$. This yields the (estimated) independent components by
\begin{equation}
    \label{invICA}
    s = Wx.
\end{equation}
Here, $W$ is an $m \times m$ "separating matrix". An essential assumption of ICA is independence across the latent components. As noted by \cite{hyvarinen2000independent}, this assumption is not unrealistic in applications and does not need to be precisely true in practice. Furthermore, we assume the latent components are not following Gaussian distributions (for more details, \cite{hyvarinen2000independent, bugli2007comparison}). The latter phenomenon concerns the orthogonal mixing matrix that can not be estimated for Gaussian variables. It is related to the property that uncorrelated joint Gaussian variables are necessarily independent. As noted in \cite{bugli2007comparison}, given the case of the Gaussian independent component, one can estimate the ICA model only up to orthogonal transformation. 

Practically, an ICA algorithm provides the matrix $W$ such that the signal estimates $s$ are as (stochastically) independent as possible, given the information in the data matrix $X$. In this study, we adopt the Fast-ICA algorithm in order to obtain our numerical results. This algorithm utilizes two (useful) pre-processing steps. The first indispensable step is to center the data at hand, i.e., to subtract the (arithmetic) column mean from each column in  $X$. By this, we obtain empirically centered (zero-mean) features. The second step is the so-called "whitening" step. Here, the goal is (following the centering step) to linearly transform each column in $X$ such that the empirical covariance matrix of $X$ after whitening equals the identity (for more details, we refer to Section 5 in \cite{hyvarinen2000independent}). The standard way to do so is by applying the eigenvalue decomposition, i.e., the empirical covariance matrix is represented with respect to its eigenvalues and eigenvectors. That representation itself is achieved with a PCA method. In this study, we employ  the function "FastICA" in \cite{scikit-learn}, where the two aforementioned steps are integrated by default.

\section{IC-Trees}
\label{sec:ictrees}

JPTs have been demonstrated to be an excellent model-free (in a machine-learning context, meaning distribution-free) way of learning the structure and parameters of shallow deterministic, probabilistic circuits. This framework consists of a binary partitioning tree, similar to decision trees in a sense given in \cite{quinlan1993program}. In the leaves, the fully factorized distributions (cf. Equation (\ref{eq:prob_sum})) consist of quantile parameterized distributions (QPDs) for each numerical variable and frequency distributions for each symbolic variable. QPDs were introduced by \cite{keelin2011quantile}. The symbolic variables present nominal outcome variables in a multinomial regression (in the context of generalized linear models), i.e., those variables are categorical and unordered. Furthermore, our approach is limited to numerical (data) features, i.e., each (random) variable takes (only) numerical values (real numbers). 
In the context of machine-learning modeling, the JPT approach to PCs results in an easily interpreted, powerful, and efficient model. However, JPTs suffer from limited predictive power, as any probabilistic circuit does. In this study, our goal is to enhance the performance of the JPTs structure by incorporating the resulting (independent) ICA vectors.

\subsection{Transformed splits}\label{sec:transsplits}
 The first enhancement of our framework, in contrast to JPTs, is to increase the expressiveness of a deterministic PC by splitting on a linear combination of the underlying features (not only on a single variable).
 Deterministic in the context of PCs (see \cite{choi2020probabilistic}) means that in a sum, at most, one summand is greater than zero for all observation units (namely a row-wise input in our data matrix).
 That property is desirable for PCs so as to be able to split on more sophisticated criteria than single variable splits. Therefore, we allow PCs to split on the domain variables' linear combinations (the resulting ICA vectors). 
 Another benefit of deterministic PCs is that they can be maximized, the so-called "maximizer circuit" (for more details, see Definition 28 in \cite{choi2020probabilistic}). 

 In our study, we employ the tree-based methods in order to split the input data into disjoint partitions ($K := |\Lambda|$, where $K$ is the number of partitions or regions), and we denote these regions by $(R_{k})_{1 \leq k \leq K}$. 
 As noted in \cite{nyga2022jpt}, the JPT approach is not limited solely to numerical data.
However, our improvement only concerns numerical variables (data features); hence we omit the notations for symbolic variables. We follow the general notation in Chapter 9.2 in \cite{hastie2009elements} for the tree-based methods. Letting $x_i = (x_{i1}, \ldots, x_{im})$ denote the $i$-th row in the data matrix $X$, which is the data vector for the $i$-th observational unit ($1 \leq i \leq n$), it holds that
\begin{align}
\label{eq:linear_decision}
    R_{1} &= \{ x_i: \sum_j^m a_{j} (x_{ij} - \bar{x}_j)< b \}, \nonumber \\ 
    \quad R_{2} &= \{ x_i: \sum_j^m a_{j} (x_{ij} - \bar{x}_j)\geq b \},
\end{align}
which we rewrite as 
\begin{align*}
    R_{1} &= \{x_i: \sum_j^m a_{j} x_{ij} < b + \sum_j^m a_{j} \bar{x}_j \}, \nonumber \\ 
    \quad R_{2} &= \{ x_i: \sum_j^m a_{j} x_{ij}  \geq b + \sum_j^m a_{j} \bar{x}_j\},
\end{align*}

where $\bar{x}_j$ is the empirical mean for the $j$-th feature, $j \in \{1, \ldots, m\}$, and $b, a_1, \ldots, a_m$ are the parameters which define the separating hyperplane.
 Hence, these hyperplanes in our application present a multivariate line separator (since we have a binary decision rule). 

The deterministic (partition) property of JPTs is preserved under linear splits since the sum in Equation \eqref{eq:linear_decision} evaluates to either less than $b$ or not less than $b$ for any observational unit, i.e., for every $x_{i}$, $i \in \{1, \ldots, n\}$.

\subsection{Transformed leaves}
\label{sec:leaf}
The joint probability distribution of a single transformed leaf is given by

\begin{align}
\label{eq:linear:pdf}
P(x_{1}, \ldots ,x_{m} | \lambda) = \frac{1}{|det(A_\lambda)|}\prod_{j}^{m} P_{\lambda_j}(e_{\lambda_j}) ,
\end{align}

where\\

\begin{align}
e_{\lambda_{j'}} = \sum_j^m a_{\lambda_{j'j}} (x_j - \bar{x}_{\lambda_{j}}).
\end{align}
Here, $\bar{x}_{\lambda_{j}}$ is the mean of the j-th variable in leaf $\lambda$, which has to be subtracted since a useful step in the ICA method is to center the data at hand (for more details, see Section \ref{sec:ica}). Furthermore, for a fixed $j' \in \{1, \ldots, m \}$ denotes the $j'$-th component of the transformation and $a_{j'j}$ is the element of the matrix $A_\lambda$, where the latter is defined  as
\begin{align}
    A_\lambda &= \begin{pmatrix}
            a_{\lambda 11} & \ldots & a_{\lambda m1} \\
            \vdots & \ddots & \vdots \\
            a_{\lambda 1m} & \ldots & a_{\lambda mm}
         \end{pmatrix}.
\end{align}
$A_\lambda$ is the Jacobian matrix of the transformation given by the matrix-vector form of the transformation for leaf $\lambda$. Since the ICA is a location and scale transformation, then the distortion due to its transformation needs to be inverted by dividing by its determinant (\cite{klenke2013probability}).


The joint probability distribution of an entire IC-Tree is expressed as
\begin{align}
\label{eq:icajpt}
    P(x_1, \ldots ,x_m) = \sum_{\lambda \in \Lambda} P(\lambda) P_\lambda(x_1, \ldots ,x_m | \lambda).
\end{align}
Here, $\Lambda$ is the set of all leaves for an individual tree, and for $\lambda \in \Lambda$ corresponds to a leaf. For convenience of notation, we refer only to a single tree and a single leaf \textemdash and not to all trees and all leaves. We, therefore, intentionally do not index trees and leaves. $P(\lambda)$ is the probability that an observational unit is assigned to a leaf, i.e., the percentage of observational units assigned to that partition leaf (for $ \lambda \in \Lambda$) during the training process.
$P_{\lambda_j}(e_{\lambda_j})$ is the univariate probability distribution of the $j$-th feature, and for $j \in \{1, \ldots, m \}$ in the transformed space for leaf $\lambda$. 
$P_{\lambda_j}(e_{\lambda_j})$ is represented by a QPD over the realizations of $e_{\lambda_j}$, where $e_{\lambda_j}$ corresponds to the linear combination of the original dimension. 

\subsection{Learning}

Learning the structure and parameters of IC-Trees can be executed efficiently by using a modified version of the C4.5 algorithm in \cite{quinlan1993program}.
In each induction step of C4.5, the ICA technique is computed and applied to the data of the current iteration. The optimal split is then calculated by minimizing the entropy and variance on the transformed axis (cf. \cite{nyga2022jpt}).
For the optimal split, the criterion is to be converted into the scalar equation of a plane to be stored compactly, as seen in Equation (\ref{eq:linear_decision}). When a stopping criterion is reached, e.g., minimum partition size, maximum depth, etc., a fully factorized leaf distribution is fitted on the transformed data.

\subsubsection{Time Complexity}

As in regular decision tree learning, the time complexity depends on the number of observational units $n$ and features $m$. Following the argumentation in \cite{sani2018computational}, the dominant term for JPTs is $O(n m^2 log_2(n))$. Since the JPTs approach needs to compute the impurity/variance of each variable, it is $m$ times more complex than ordinary decision tree learning. In IC-Trees, this term is scaled by the run time of the FastICA algorithm, which is $2(n+1)m$ per iteration (cf. \cite{zarzoso2006fast}). Since the iteration is just a constant set in the hyperparameters, it can be ignored for the big O notation. This gives a total time complexity of $O(n m^2 log_2(n) 2(n+1)m)$, which is still efficient.

\subsubsection{Space Complexity}

The space required to train our framework is composed  $n$ by $m$ (i.e., the number of observational units multiplied by the number of features), the number of nodes, and the number of numeric variables. For each inner node that splits on the underlying features (denoted by m), we need to store $m + 1$ parameters. For symbolic variables, only one parameter is required. For the leaf distributions, a square matrix with $m^2$ elements needs to be stored, and another vector for the mean is needed to subtract in the ICA. In addition, the distributions need to be stored. As they are model-free, the maximum number of parameters equals the number of observational units per leaf.

\subsection{Inference}

Performing inference in shallow PCs boils down to describing inference in leaf distributions. Therefore, we will discuss inference for the query types proposed in \cite{choi2020probabilistic} for the leaf distributions described in Section \ref{sec:leaf}.

First off, exact marginal inference in IC-Trees is not tractable. Marginal queries require integration over hyperrectangles. Integrating a hyperrectangle in a transformed (leaf) coordinate system results in the integration over a parallelepiped. For ordinary areas, this is calculable.
Since QPDs are mixtures of functions, where the domain of a function is represented by indicators, integration of indicators is required. These indicators are not aligned with the bounds of the integrals. Hence the integral over every possible combination of indicators has to be calculated. The size of this calculation is exponentially large in the number of indicators per variable and, therefore, should be omitted.
Combining model-free (quantile parameterized) distributions with transformations of any (non-trivial) kind leads, in general, to a distribution which does not belong to the family of stable distributions (cf. \cite{borak2005stable}).

\subsubsection{Sampling}
Since exact marginal inference is not tractable, most inferences have to be designed on the basis of sampling. Sampling QPDs is computationally efficient and can be considered as two-stage sampling, where a discrete random variable is sampled first.

QPDs are defined in terms of countably many, mutually non-overlapping and exhaustive intervals covering the domain of a continuous random variable, i.e., half-open intervals on $\mathbb{R}$ (see, in \cite{driver2020probability} and \cite{keelin2011quantile}).
The index set enumerating these intervals can be re-expressed as the domain of a discrete random variable. The probability of the state describing an interval $[a, b)$ is given by the integral from $a$ to $b$ of the function in the interval. 

In the first stage, a state (interval) of this random variable has to be sampled.
In the second stage, a sample from the distribution of the sample interval has to be drawn. As the basis distributions of QPDs in JPTs and IC-Trees are uniform distributions, sampling is easily executed.

However, the sampling procedure can produce samples that are inconsistent with the path of a leaf. This is due to the fact that the coordinate system from which the hyperplane of a split in the path originates is (potentially) different from the coordinate system of a leaf.

For instance, a tree could split with the criterion $-0.5x - 1.3y = -3.4$ and a leaf models the points using the transformation $A_1 = \begin{pmatrix}
    3.8 & 1.3 \\
    42.9 & -10.2 
\end{pmatrix}$. Since the variables are modeled independently, the constraint from the path cannot be modeled accurately.

Therefore, samples that are inconsistent with the path either have to be discarded or re-sampled. Yet, the number of such samples is, by definition, small since the splits are distance-maximizing and thus drawn through empty regions in the original data. 

Sampling the whole tree is conducted in a three-stage sampling procedure, where first, a leaf $\lambda$ is sampled from the discrete distribution over $\Lambda$ (i.e., $P(\Lambda)$), and then the two stages described above are applied to the sampled leaves.
\subsubsection{Evidence based inference}

Creating conditional distributions in IC-Trees is not as straightforward as in PCs. Since the evidence is defined in the original dimensions and not the transformed dimensions, the distributions cannot simply be restricted. Evidence has to be applied approximately prior to the sampling process. The hyperrectangle that defines the evidence in the original dimensions converts to a parallelepiped in the leaf dimensions. The distributions in the leaf can be restricted to the intervals that completely subsume the parallelepiped. Hence, the samples are more likely to be consistent with the evidence, but this is not guaranteed. 

The \textbf{marginal} probability is then calculated by counting the number of samples consistent with the marginal query and dividing that number by the total amount of samples.

The \textbf{moments} of the original random variables are calculated by sampling from the leaf distributions, transforming the samples back to the original space, and calculating the empirical moment of the samples, e.g., empirical mean, variance, etc.

\textbf{Most probable explanations} can be calculated in the dimensions of the leaves. Due to the definition of QPDs in JPTs, the most probable explanations are not single points but whole areas. However, those areas with their transformation can be used to answer the most probable explanations. Alternatively, one can sample from this distribution in leaf dimensions and transform the points to the original domain to get precise answers.

Inference in the entire IC-Tree is straightforward as the tree is a deterministic linear combination of its leaf distributions.

\subsection{Symbolic information}

The use of symbolic variables does not change in IC-Trees, since no general algebra can be imposed on them. Therefore, no transformations or calculations can be done on them. They are therefore treated in the same way as in JPTs (cf. \cite{nyga2022jpt}).

\section{Experiments}
\label{sec:experiments}

This section provides empirical evidence that our approach outperforms and increases expressiveness over the classical counterpart, i.e., the JPTs framework (cf. \cite{nyga2022jpt}).

The commonly used (numerical) metric to evaluate the quantitative performance proposed in the literature (see, e.g., in \cite{nyga2022jpt}, \cite{peharz2020einsum} and \cite{trapp2020thesis}), is the average log-likelihood in Equation \eqref{eq:avg_ll}. Namely, the latter metric can be evaluated as 
\begin{equation}
    \frac{1}{n}\sum_{i=1}^n L_{1, \ldots, m}(x_{i1}, \ldots, x_{im})\label{eq:avg_ll},
\end{equation}
where $L_{1, \ldots, m}$ denotes the joint log-likelihood function of the $m$ random variables (features) in the data set. For calculating $L_{1, \ldots, m}$, we used the logarithmic form of Equation \eqref{eq:icajpt}. As in Section \ref{sec:transsplits}, $(x_{i1}, \ldots, x_{im})$ denotes the data (row) vector for the $i$-th observational unit ($1 \leq i \leq n$) in Equation \eqref{eq:avg_ll}. 

\begin{figure*}
    \centering
    \includegraphics[width=\textwidth]{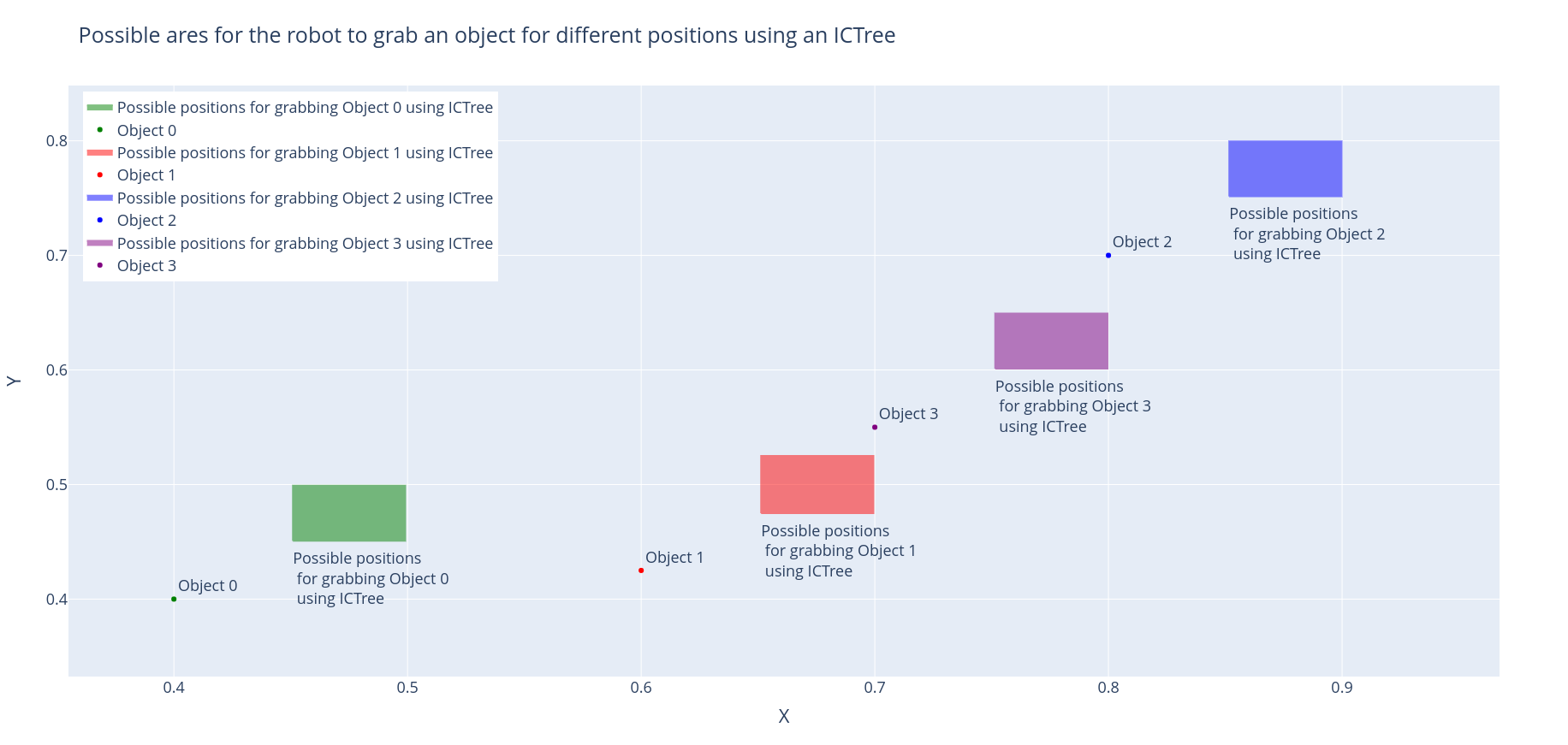}
    \caption{Visualization of the picking task specified in Section \ref{sec:intro}. 
    The dots describe an object in the 2D plane. The rectangle with the same color as a dot describes the possible area for a robot to grab the selected object inferred by an IC-Tree with 32 parameters.}
    \label{fig:ict_predictions_grabbing}
\end{figure*}

First off, we revisit the example of the introduction, where a robot has to grab an object from the upper right corner. In contrast to the predictions performed by JPTs (Figure \ref{fig:jpt_predictions_grabbing}), the predictions made by IC-Trees are shown in Figure \ref{fig:ict_predictions_grabbing}. In this figure, multiple (random) objects are simulated, and the model is queried for all maximum likely positions where the robot can reach the object. As intended by the task specification, the possible areas are always the upper right area concerning the queried object. In our approach, only 32 parameters and one leaf are used. 
The JPTs approach achieved an average log-likelihood of $1.87$, whereas the IC-Trees achieved an average log-likelihood of $5.96$.
Therefore, the desired effect to describe the dependencies has been reached. Through this experiment, we illustrate the benefit of our approach over the JPTs framework, i.e., in a (highly) dependent (robotics) environment, one can significantly profit from our approach.

Another experiment involves a simulated robot performing tasks in the cognitive robot abstract machine (CRAM) framework (cf. \cite{koralewski2019self}). Here, a robot did 1015 trials of a handover task.
In the handover task, a robot has to pick up an object with one hand, pass it over to the other hand, and drop it at a location.
In every experiment trial, the robot recorded seven symbolic variables and 18 continuous variables.
A JPT with 2219 parameters has been fitted. An average log-likelihood of $67.86$ was achieved. In comparison, IC-Trees with 4097 parameters were fitted. Our approach achieved an average log-likelihood of $105.52$. 

In Table \ref{tab:empirical}, the results of a large empirical evaluation of IC-Trees on eight benchmark data sets taken from the UCI machine learning repository are reported (see \cite{Dua:2019}). In all experiments, 10\% of the data was randomly selected as a test set. No (data) pre-processing steps were applied to these data sets.
Overall, as one can see from the empirical evaluation by (\cite{nyga2022jpt}) (cf. Table 3 in this reference), our methodology outperforms the classical JPTs. IC-Trees achieved a higher average log-likelihood on the test set in $29$ out of $39$ comparable experiments. Another interesting observation is that the number of parameters used in the IC-Trees was lower in $24$ out of $42$ experiments than the number of parameters used in JPTs (\cite{nyga2022jpt}). Even though the transformations used in IC-Trees require parameters, the overall parameter count was low in many experiments. This is explained by the simplicity that the data may exhibit in the transformed dimensions. Hence, the QPDs require fewer parameters to approximate the distribution. However, one has to keep in mind that exact marginal inference is not possible in IC-Trees. Unfortunately, comparisons to other studies (i.e., \cite{peharz2020einsum}, \cite{trapp2020thesis}) are not applicable since they used binary data sets. 


\section{Related Work}
\label{sec:related_work}
Combining linear models with tree-based models was explored recently for discriminative modeling (see, \cite{zhang2019regression}, \cite{ilic2021explainable}).
However, their approaches neither discussed linear splits nor a full probabilistic context. 
 Recently, the applications of transformations in probability distributions have been discussed (see, (\cite{driver2020probability} and \cite{siegrist2017probability}). To this end, such studies have paved the way for learning transformed distributions.
Assuming a set of latent variables that are independent is richly described by \cite{bugli2007comparison}. However, to the best of our knowledge, we are the first authors to integrate distribution-free models with the ICA. 
PCs provide the general framework to describe and discuss probabilistic modeling. They are discussed comprehensively in \cite{choi2020probabilistic}. 
JPTs provide ideas to learn the structure and parameters for model-free deterministic PCs. However, \cite{nyga2022jpt} do not present new ideas for computation nodes in PCs. 


\section{Conclusion and Outlook}
\label{sec:conclusion}
In this study, we have introduced a framework entitled independent component analysis - joint probability trees (IC-Trees), which is an extension of JPTs. In general, our approach consists of two steps. First, it utilizes the ICA methodology to enhance the expressive power of PCs \textemdash by incorporating the dependency among the (data) features in PCs. Second, our framework exploits the JPTs approach to deterministic PCs with the ICA procedure. Throughout the paper, we discussed the complexity, (storage) space, and inferential limitations of our approach. Furthermore, we support our approach by providing empirical results from several real-data examples (including two examples in robotics) and illustrating its benefits over the JPTs-based counterpart.

There are several potential aspects for further research. First, it might be of interest to explore the usage of stable distributions to overcome inferential limitations. Second, it might be captivating to enhance the numerical stability of the ICA by selecting the number of latent components, for example based on an information-theoretic criterion. Finally, it would be interesting to utilize the concepts for general PCs (i.e., not necessarily a partitioning tree) by introducing product nodes as soon as some parts of the covariance matrix reach identity.
\begin{table*}[h]
\caption{Results of the evaluation of IC-Trees on eight benchmark data sets of the UCI machine learning repository (cf. \cite{Dua:2019}) for different hyperparameters. "Min samples per leaf" means that at least the respective percentage of data points available for training must be represented by any leaf of the tree. "Min samples per leaf" = 90\%, thus 
resulting in an IC-Trees framework with only one transformed leaf, i. e. a set of independent distributions over all transformed variables. ``0-likelihood test samples'' determines the percentage of test samples with 0 likelihood. 
This may happen to samples lying outside the convex hull of the 
training data, where 0 probability mass is assigned by the 
QPDs. In all experiments, the number of iterations for FastICA was 1000. We set the number of components equal to the number of variables.}
\resizebox{.935\textwidth}{!}{
 \begin{tabular}{p{5cm}|c|c|c|c|c}
     \toprule
      \textbf{Dataset} & \textbf{Min samples per leaf} & \textbf{Model Size} &  \textbf{Average Train Log-Likelihood} & \textbf{Average Test Log-Likelihood} &  \textbf{0-likelihood Test Samples} \\
     \hline\hline
    \multirow{3}{*}{\shortstack[l]{\textbf{IRIS Dataset}\\Observational Units:  150\\Variables:  5}} 
    & 90\% & 71 & -2.69 & -2.94 & - \\
    & 40\% & 175 & -1.04 & -1.64 & 13\% \\
    & 20\% & 268 & 0.15 & -2.18 & 13\% \\
    & 10\% & 497 & 0.72 & -1.34 & 53\% \\
    & 5\% & 915 & 3.95 & 0.78 & 87\% \\
    & 1\% & 3562 & 16.19 & - & 100\% \\\hline
    \multirow{3}{*}{\shortstack[l]{\textbf{Adult Dataset}\\Observational Units:  32561\\Variables:  15}}
    & 90\% & 211 & -44.39 & -44.48 & - \\
    & 40\% & 432 & -42.59 & -42.68 & - \\
    & 20\% & 877 & -40.42 & -40.55 & - \\
    & 10\% & 1764 & -40.17 & -40.25 & 1\% \\
    & 5\% & 2853 & -40.61 & -40.7 & 6\% \\
    & 1\% & 15370 & -41.27 & -45.25 & 58\% \\\hline
    \multirow{3}{*}{\shortstack[l]{\textbf{Dry Bean Dataset}\\Observational Units:  13611\\Variables:  17}}
    & 90\% & 470 & 23.02 &  22.72 & - \\
    & 40\% & 948 & 23.56 & 23.62 & - \\
    & 20\% & 1418 & 31.57 & 31.48 & - \\
    & 10\% & 3282 & 44.06 & 43.63 & - \\
    & 5\% & 6552 & 45.67 & 45.25 & 2\% \\
    & 1\% & 36700 & 55.01 & 53.39 & 18\% \\\hline
    \multirow{3}{*}{\shortstack[l]{\textbf{Wine Dataset}\\Observational Units:  178\\Variables:  13}}
    & 90\% & 330 & -16.76 &  -18.34 & 11\% \\
    & 40\% & 767 & -13.22 & -17.66 & 28\% \\
    & 20\% & 1615 & -8.36 & -18.26 & 67\% \\
    & 10\% & 2528 & -2.87 & - & 100\% \\
    & 5\% & 4408 & 10.77 & - & 100\% \\
    & 1\% & 31963 & 63.10 & - & 100\% \\\hline
    \multirow{3}{*}{\shortstack[l]{\textbf{Wine Quality Dataset}\\Observational Units:  6497\\Variables:  13 }} 
    & 90\% & 290 & -5.36 &  -5.4 & - \\
    & 40\% & 587 & -5.00 &  -5.0 & 1\% \\
    & 20\% & 1158 & -3.12 & -3.12 & 2\% \\
    & 10\% &  2005 & -2.78 & -2.93 & 2\% \\
    & 5\% & 4096 & -2.14 & -2.92 & 4\% \\
    & 1\% & 29238 & 3.05 &  -2.0 & 28\% \\\hline
    \multirow{3}{*}{\shortstack[l]{\textbf{Bank and Marketing Dataset}\\Observational Units:  45211\\Variables:  17}}
    & 90\% & 189 & -34.47 & -35.66 & - \\
    & 40\% & 381 & -36.61 &  -36.71 & - \\
    & 20\% & 747 & -35.36 & -35.66 & - \\
    & 10\% & 1465 & -41.83 & -42.12 & 81\% \\
    & 5\% & 2084 & -35.48 & -35.74 & 66\% \\
    & 1\% & 5948 & -35.96 & -36.56 & 79\% \\\hline
    \multirow{3}{*}{\shortstack[l]{\textbf{Raisin Dataset}\\Observational Units:  900\\Variables:  8 }} 
    & 90\% & 137 & -29.65 & -29.65 & 6\% \\
    & 40\% & 265 & -27.69 & -27.71 & 9\% \\
    & 20\% & 565 & -27.18 &  -27.40 & 12\% \\
    & 10\% & 1299 & -25.94 & -27.63 & 17\% \\
    & 5\% & 2736 & -23.73 &  -26.80 & 0.23 \\
    & 1\% & 9082 & -10.73 & -28.26 & 96\% \\\hline
    \multirow{3}{*}{\shortstack[l]{\textbf{Abalone Dataset}\\Observational Units:  4177\\Variables:  9}}
    & 90\% & 168 & 9.96 & 9.82 & 1\% \\
    & 40\% & 333 & 10.61 & 10.35 & 1\% \\
    & 20\% & 494 & 10.71 & 10.43 & 1\% \\
    & 10\% & 1484 & 11.37 & 10.8 & 3\% \\
    & 5\% & 2703 & 11.73 & 10.81 & 6\% \\
    & 1\% & 16597 & 14.99 & 10.59 & 39\% \\\hline
\end{tabular}}
\label{tab:empirical}
\end{table*}
\FloatBarrier

\bibliographystyle{unsrtnat}
\bibliography{references}  

\section*{Supplementary Material}
\subsection*{Visual Model Explanation}
\begin{figure}
    \centering
    \includegraphics[width=\textwidth]{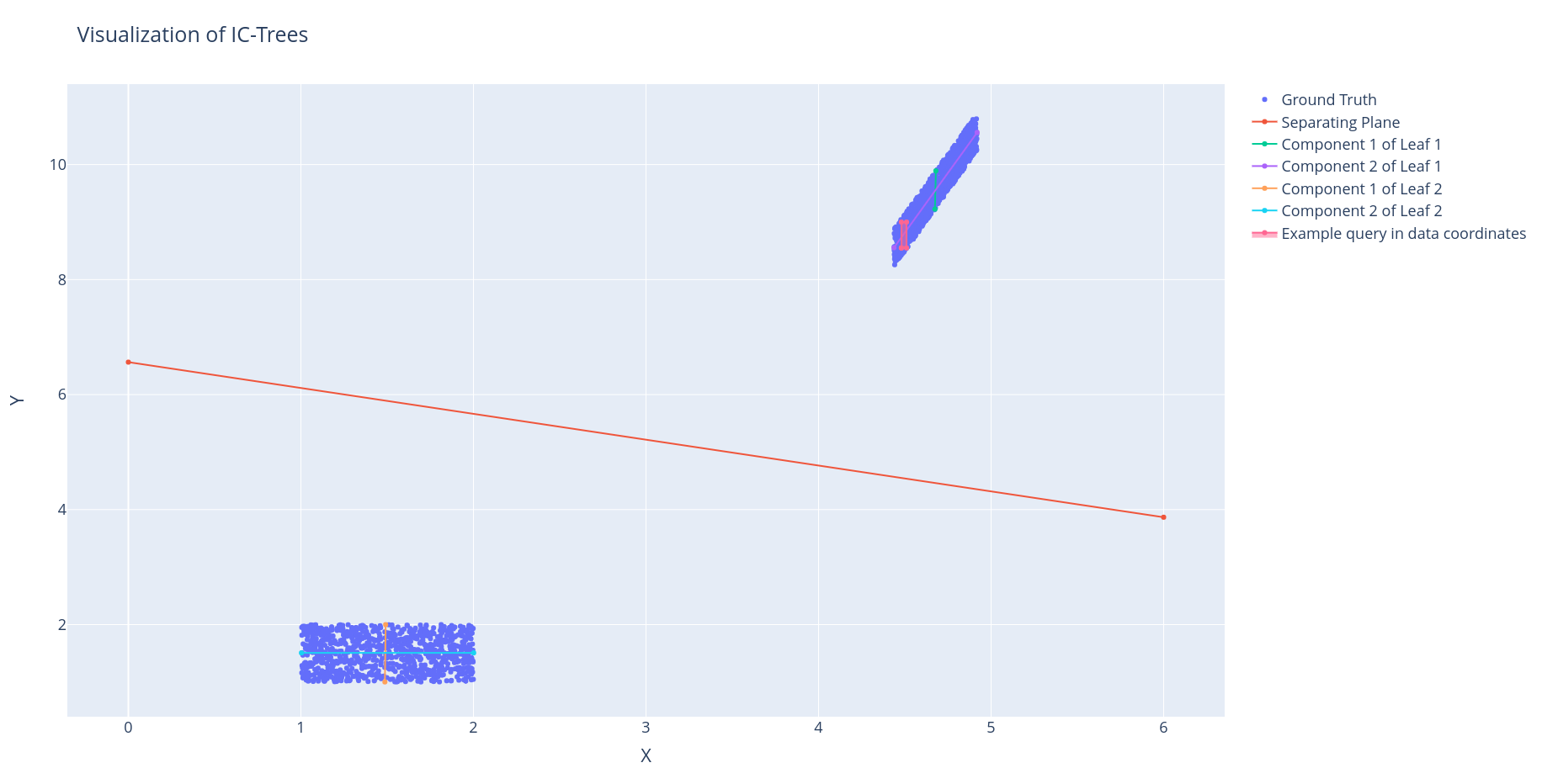}
    \caption{Density of points sampled from two Uniform distributions in 2D approximated by an IC-Tree. The lower left cluster is created by combining two independent uniform distributions. The upper right points are samples from two linear dependent uniform distributions. The red line visualizes the learned hyperplane that separates the leaf distributions. The lines inside the clusters show the directions of the transformed components.}
    \label{fig:two_uniforms}
\end{figure}

\begin{figure}
    \centering
    \includegraphics[width=\textwidth]{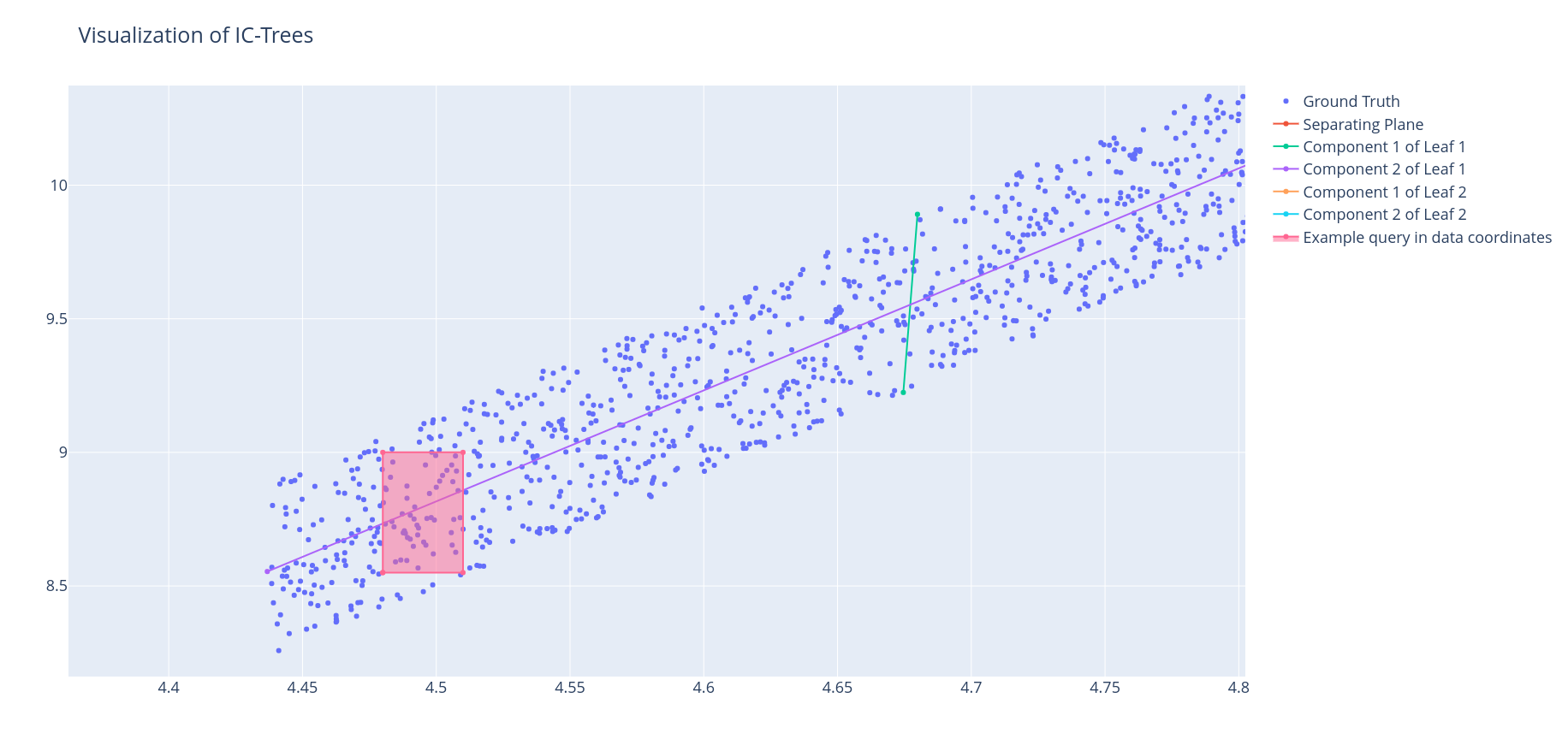}
    \caption{Closer visualization of the upper right cluster from Figure \ref{fig:two_uniforms}. The red rectangle visualizes a marginal query $P(4.48 \leq X < 4.51, 8.55 \leq Y < 9)$ in the dimensions of the original variables.}
    \label{fig:query_c2_data}
\end{figure}

\begin{figure}
    \centering
    \includegraphics[width=\textwidth]{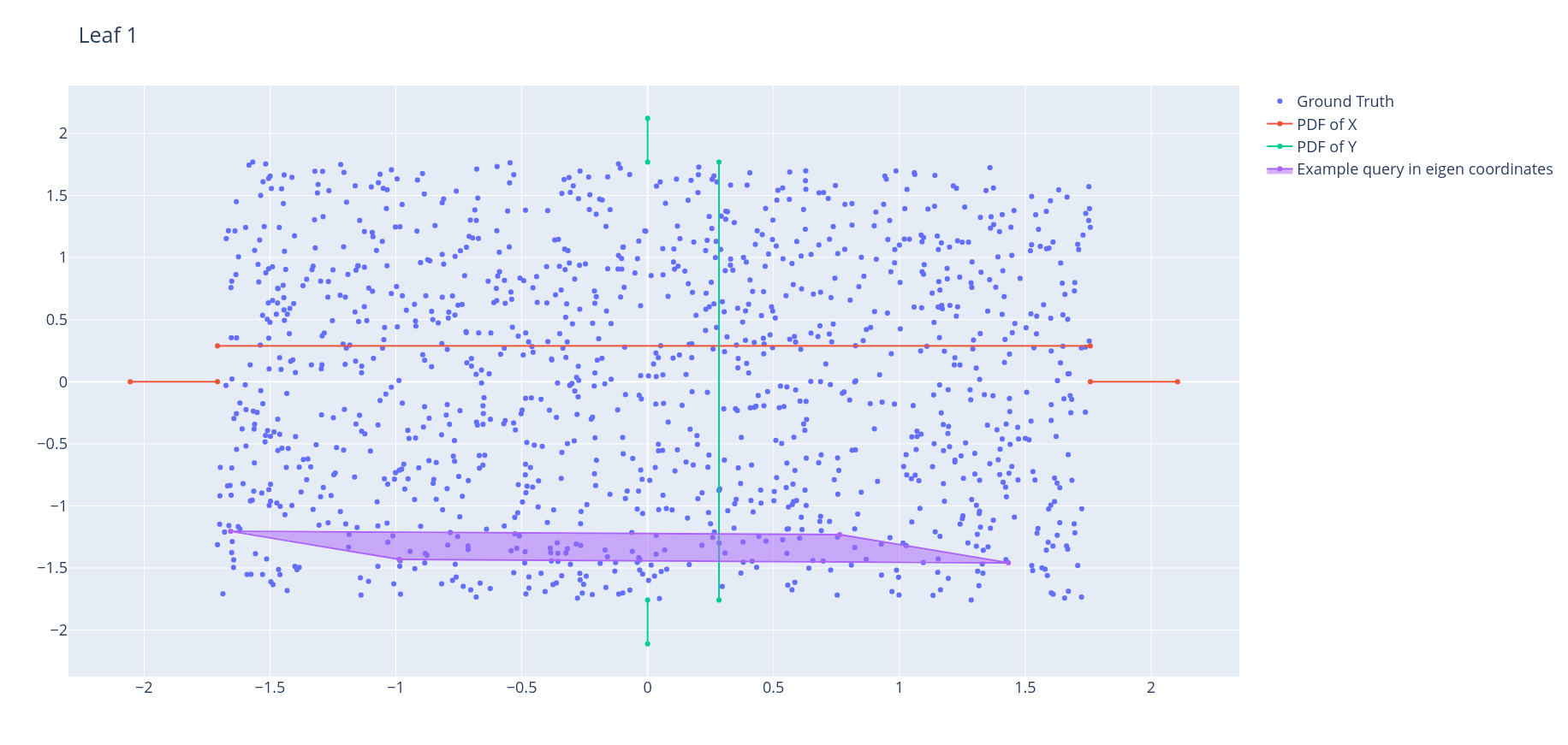}
    \caption{Closer visualization of the upper right cluster from Figure \ref{fig:two_uniforms}. This time the transformed dimensions of the corresponding leaf are displayed. It can be confirmed visually that the new dimensions are (almost) perfectly independent. The purple rectangle visualizes the marginal query from Figure \ref{fig:query_c2_data} in the dimensions of the leaf variables. As explained in the main article, the query transforms to a parallelogram. The green and red lines display the uniform distributions that are fitted on the new axis.}
    \label{fig:query_c2_eigen}
\end{figure}

\begin{figure}
    \centering
    \includegraphics[width=\textwidth]{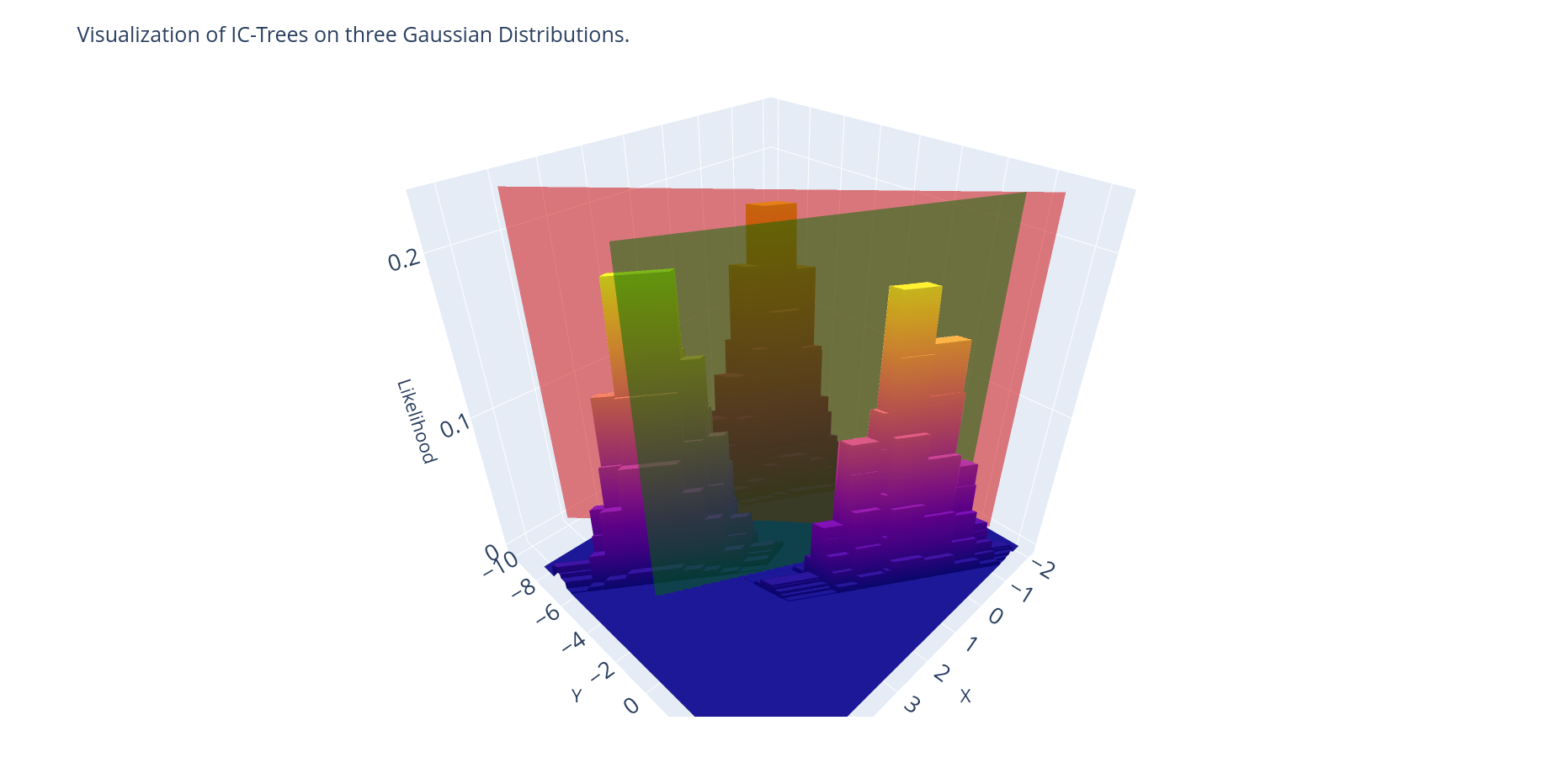}
    \caption{Density of points sampled from three Gaussian distributions approximated by an IC-Tree. The green and red surfaces describe the hyperplanes that are created as a splitting criterion during training. The parallelepiped structure that is generated by the QPDs on the transformed dimensions is clearly visible.}
    \label{fig:three_gaussians}
\end{figure}

The purpose of this section is to supply further materials that visually support the understanding of the transformations used in IC-Trees. The Figures \ref{fig:two_uniforms}, \ref{fig:query_c2_data}, \ref{fig:query_c2_eigen} and \ref{fig:three_gaussians} visualize the structures that are used in IC-Trees. 

\end{document}